\documentclass[letterpaper]{article} 
\usepackage{aaai2026new}  
\usepackage{times}  
\usepackage{helvet}  
\usepackage{courier}  
\usepackage[hyphens]{url}  
\usepackage{graphicx} 
\usepackage{natbib}  
\usepackage{caption} 
\newif\ifappendix
\appendixtrue
\usepackage[utf8]{inputenc}

\usepackage{subcaption}
\usepackage{amsmath}
\usepackage{amsthm}
\usepackage{booktabs}
\usepackage{algorithm}
\usepackage[noend]{algorithmic}
\usepackage{amssymb}
\usepackage{complexity}
\usepackage{thm-restate}
\usepackage{todonotes}
\usepackage{tikz,pgf}
\usepackage{array}
\definecolor{AssumpBlue}{RGB}{82,188,224}   
\definecolor{RulePurple}{RGB}{214,162,215}  
\usetikzlibrary{shapes,backgrounds,shapes.geometric,shadows,patterns}
\usetikzlibrary{positioning}
\tikzstyle{arg}=[draw,circle,inner sep=1pt,minimum size=.5cm]
\tikzstyle{rarg}=[draw=RulePurple,fill=RulePurple,circle,inner sep=1pt,minimum size=.5cm]
\tikzstyle{aarg}=[draw=AssumpBlue,fill=AssumpBlue,circle,inner sep=1pt,minimum size=.5cm]
\usepackage[dvipsnames]{xcolor}

\let\S\defaultS

\usepackage{etoolbox}
\usepackage{thm-restate}
\usepackage{todonotes}
\newcommand{\TODO}[1]{}

\newtheorem{theorem}{Theorem}[section]

\newtheorem{proposition}[theorem]{Proposition}

\newtheorem{observation}[theorem]{Observation}

\newtheorem{definition}[theorem]{Definition}

\newtheorem{example}[theorem]{Example}

\newcommand{\contrary}[1]{\overline{#1}}
\newcommand{\contraryempty}{\contrary{\phantom{a}}}

\newcommand{\lit}{\ensuremath{\mathcal{L}}}
\newcommand{\rules}{\ensuremath{\mathcal{R}}}
\newcommand{\asm}{\ensuremath{\mathcal{A}}}
\newcommand{\contraryc}[1]{\overline{#1}}

\newcommand{\stable}{{\mathit{stb}}}
\newcommand{\stb}{\stable}

\newcommand{\cf}{\mathit{cf}}

\newcommand{\la}{\leftarrow}

\newcommand{\dependency}{\mathit{G}}
\newcommand{\body}{\mathit{body}}
\newcommand{\head}{\mathit{head}}

\usepackage{cancel}

\newcommand{\ie}{i.e.}

\newcommand{\wrt}{w.r.t.}
\newcommand{\suth}{s.t.\, }

\newcommand{\tuple}[1]{\ensuremath{\left(#1\right)}}

\renewcommand{\D}{\ensuremath{\mathcal{D}}}

\usepackage{multirow}
\usepackage{multicol}
\usepackage{newfloat}
\usepackage{listings}
\DeclareCaptionStyle{ruled}{labelfont=normalfont,labelsep=colon,strut=off} 
\floatstyle{ruled}
\newfloat{listing}{tb}{lst}{}
\floatname{listing}{Listing}

\title{Heterogeneous Graph Neural Networks for Assumption-Based Argumentation}
\author{
    Preesha Gehlot\textsuperscript{\rm 1}\equalcontrib,
    Anna Rapberger\textsuperscript{\rm 1,2}\equalcontrib,
    Fabrizio Russo\textsuperscript{\rm 1}\equalcontrib,
    Francesca Toni\textsuperscript{\rm 1}
}
\affiliations{
    \textsuperscript{\rm 1}Imperial College London, Department of Computing\\
    \textsuperscript{\rm 2}Technical University Dortmund, Faculty of Computer Science\\
    \{a.rapberger, fabrizio\}@imperial.ac.uk
}

\usepackage{bibentry}

\begin{document}

\maketitle

\begin{abstract}
Assumption‐Based Argumentation (ABA) is a powerful structured argumentation formalism, but exact computation of extensions under stable semantics is intractable for large frameworks.  We present the first Graph Neural Network (GNN) approach to approximate credulous acceptance in ABA.  
To leverage GNNs, we model ABA frameworks via a dependency graph representation encoding assumptions, claims and rules as nodes, with heterogeneous edge labels distinguishing support, derive and attack relations.  
We propose two GNN architectures—ABAGCN and ABAGAT—that stack residual heterogeneous convolution or attention layers, respectively, to learn node embeddings.  Our models are trained on the ICCMA~2023 benchmark, augmented with synthetic ABAFs, with hyperparameters optimised via Bayesian search.  Empirically, both ABAGCN and ABAGAT outperform a state‐of‐the‐art GNN baseline that we adapt from the abstract argumentation literature, achieving a node‐level F1 score of up to 0.71 on the ICCMA instances. Finally, we develop a sound polynomial time extension‐reconstruction algorithm driven by our predictor: it reconstructs stable extensions with F1 above 0.85 on small ABAFs and maintains an F1 of about 0.58 on large frameworks. Our work opens new avenues for scalable approximate reasoning in structured argumentation.
\end{abstract}

\section{Introduction}

Computational argumentation provides formal tools for modelling defeasible reasoning over conflicting information.  In the \emph{abstract} setting, arguments are treated as opaque 
entities and asymmetric conflicts between them, so-called attacks, as a binary relation, giving rise to abstract argumentation frameworks (AFs)~\cite{Dung95}, which can also be understood as graphs (with arguments as nodes and attacks as edges).  By contrast, \emph{structured} formalisms expose the internal composition of arguments.  \emph{Assumption‐Based Argumentation (ABA)}~\cite{BondarenkoDKT97} is a prominent structured framework, enabling reasoning in domains such as decision support~\cite{vcyras2021assumption}, 
planning~\cite{Fan18a}, and causal discovery~\cite{causal}.

The building blocks of ABA are \emph{assumptions}, which are the defeasible elements of an ABA framework (ABAF), and \emph{inference rules}, which are used to construct \emph{arguments} based on 
\emph{assumptions}. \emph{Conflicts} arise between assumption sets $S$ and $T$ if the conclusion of an argument constructed from $S$ is the \emph{contrary} of some assumption in $T$; in this case we say that $S$ attacks $T$. 

\emph{Argumentation semantics} provide criteria which render sets of assumptions 
(so-called \emph{extensions}) jointly acceptable. 
One of the most popular semantics is the 
semantics of \emph{stable extensions} which accepts a set of assumptions $S$ only if it has no internal conflicts and attacks every 
$\{x\}$ with $x$ an assumption not 
in $S$.
While stable semantics neatly characterises acceptable assumption sets, they are hard to compute: verifying \emph{credulous acceptance} w.r.t.\ stable extension semantics, i.e., checking if an assumption is contained in any stable extension, is \NP-complete~\cite{credulous-acceptance}. 
Credulous acceptance plays an important role in several settings, e.g., ABA learning~\cite{AngelisPT24} relies on it to learn ABAFs from data. Since these settings require fast solutions and real-world ABA frameworks can be very large~\cite{causal}, relying on exact approaches is often not viable. 

In this work, we address this issue by using \emph{Graph Neural Networks (GNNs)} \cite{GNN} to predict 
credulous acceptance.
GNNs leverage the relationships between nodes and edges in a graph to learn representations that capture its structure and allow to predict labels for nodes, edges or the whole graph. 
Their specialised architecture allows GNNs to \textit{amortise} the graph analysis and use its stored weights to predict feature of interest (e.g. node labels, as in our case) in constant time after the initial training, thus introducing significant efficiency at inference time.
While GNNs have been successfully used to approximate the acceptability of AFs seen as graphs, e.g., as in~\cite{Kulhmann,Malqmvist,DBLP:conf/comma/CibierM24}, to the best of our knowledge
their potential remains unexplored for ABA and any other structured argumentation formalism to date.
To make GNNs applicable for ABA, we represent ABAFs via  \emph{dependency graphs}. 
In 
summary, our contributions are as follows:
\begin{itemize}
  \item We introduce a faithful dependency graph encoding of ABAFs that distinguishes assumption, claim and rule nodes as well as support, derive, and attack edges.
  \item We develop two heterogeneous GNN architectures— ABAGCN and ABAGAT—composed of residual stacks of relation‐specific convolutional or attention layers, respectively, enriched with learnable embeddings and degree‐based features.
  \item We implement a training and evaluation pipeline for credulous acceptance combining 380 ICCMA~2023 ABA benchmarks\footnote{Available at \url{https://zenodo.org/records/8348039}} 
  with 19,500 additional, synthetically generated ABAFs.
  Our experiments demonstrate that both ABAGCN and ABAGAT outperform an AF‐based GNN baseline based on mapping ABAFs onto AFs and then applying state-of-the-art AFGCNv2, achieving node‐level F1 up to 0.71 on small ABAFs and 0.65 overall.
  \item Based on our credulous acceptance predictor, we develop a sound poly‐time extension‐reconstruction algorithm which computes stable extensions with high fidelity.
  Our algorithm attains F1 of 0.85 on small frameworks and remains above 0.58 on large ABAFs.
\end{itemize}
Trained models, code to reproduce data and experiments are provided at \url{https://github.com/briziorusso/GNN4ABA}.

\section{Related Work}\label{sec:related}

\citeauthor{Kulhmann}~\citeyearpar{Kulhmann} were the first to frame credulous acceptance in an AF as a graph‐classification problem. They generated random AFs, labeled each argument’s acceptability using the exact CoQuiAAS solver~\cite{CoQuiAAS}, and trained a Graph Convolutional Network (GCN) on two feature sets: the raw adjacency matrix and adjacency augmented with each node’s in‐ and out‐degree. Including degree information improved accuracy to \(\sim\!\!80\%\), with F1 scores below 0.4, and class‐balancing during training enhanced detection of under‐represented arguments. Despite only marginal effects from graph topology or dataset size, their GCN reduced runtime from over an hour (CoQuiAAS) to under 0.5s for a full test set.

\citeauthor{Malqmvist}~\citeyearpar{Malqmvist} then proposed AFGCN, which augments node features with 64‐dimensional DeepWalk embeddings~\cite{deepwalk} before feeding them—alongside the AF’s adjacency matrix—through 4–6 GCN–dropout blocks with residual connections. To address the skewed accept/reject ratio in real AFs, they exclude AFs 
with very few accepted arguments and introduce a randomised per‐epoch masking of labels, forcing the model to infer hidden acceptabilities. This scheme, rather than network depth or explicit class‐balancing, accounts for most of AFGCN’s increase to \(\sim\!\!82\%\) accuracy on rebalanced data ($62\%$ reported for \cite{Kulhmann} on the same data).
\citeauthor{afgcnv2}~\citeyearpar{afgcnv2} enrich argument representations with graph‐level metrics (centralities, PageRank, colouring) and use four GCN–ReLU–dropout blocks plus an epoch‐wise reshuffle‐and‐rebalance routine and a pre‐check on the grounded extension~\cite{Dung95} for faster and more accurate inference, proposing AFGCNv2. 

\citeauthor{DBLP:conf/comma/CibierM24}~\citeyearpar{DBLP:conf/comma/CibierM24} optimise AFGCNv2 by reimplementing AF parsing and graph‐metric computation in Rust, slashing preprocessing time and memory usage. They evaluate AFGCNv2 against two variants—adding five semantic features to random features, or retaining only 11 ``meaningful" features—and assess each with and without dropout. Their results show that pruning degrades accuracy, and that the no‐dropout version of ``feature-enhanced" version achieves the best overall performance (\(\sim\!\!83\%\) accuracy vs \(\sim\!\!75\%\) reported for AFGCNv2 on the same data), highlighting the importance of rich semantic features and careful dropout placement. Finally, they introduce AFGAT, a three‐layer GATv2 (an updated attention mechanism \cite{brody2022gatv2} with multi‐head attention, see \S\ref{sec:GNNs}), which surpasses all GCN variants obtaining \(\sim\!\!87\%\) on the same benchmark data from ICCMA. 

\citeauthor{Craandijk}~\citeyearpar{Craandijk} propose AGNN, which learns argument embeddings through iterative message passing that internalises conflict‐freeness and defense, then uses the resulting acceptance probabilities to drive a constructive backtracking algorithm for extension enumeration. We build on this enumeration strategy for our proposed extension-reconstruction algorithms in \S\ref{sec:eval}, adapting it to the nuances of ABAFs instead of AFs.

We adopt AFGCNv2~\cite{afgcnv2}
(with AFs instantiated with ABAFs) as our baseline
\footnote{Available at \url{https://github.com/lmlearning/AFGraphLib/tree/main/AFGCNv2}. We could not use the versions from \cite{DBLP:conf/comma/CibierM24} as no weights were available at the time of writing.} but, inspired by the results in \cite{DBLP:conf/comma/CibierM24}, we replace its handcrafted features with learnable embeddings and experiment with both GCNs and GAT layers. Additionally, we handle class imbalance via loss weighting rather than mini‐batch rebalancing and disable the grounded‐extension heuristic to test the models' predictive accuracy on the entire frameworks rather than only on the non-grounded portion of it.
In contrast to the AF‐focused solvers discussed, we present the first native approximate solver for ABAFs, combining an enriched 
graphical representation with heterogeneous GCN and GAT‐based prediction of assumption acceptability incorporated in an AGNN‐inspired extension-reconstruction algorithm.  

\section{Preliminaries}
\label{sec:background}
We recall the relevant elements of abstract and assumption-based argumentation, as well as GNNs.
\subsection{Computational Argumentation}
An argumentation framework (AF) \cite{Dung95} is a directed graph $F \!=\! (\A,\R)$
where $\A$ are arguments and $\R\!\subseteq \! \A\times \A$ is an  \textit{attack} relation. 
For $x,y\!\in \!\A$, if $(x,y)\!\in\! \R$ we say $x$ \textit{attacks} $y$;
for $S,T\subseteq \A$,
if
$(x,y)\!\in \!\R$ for some $x \!\in \! A, y \!\in\! T$, we say $S$ \textit{attacks} $T$;
$E$ is \emph{conflict-free} ($E\in \cf(F)$) iff it does not attack itself.
A semantics $\sigma$ is a function that assigns to each AF a set of sets of arguments, so-called \emph{extensions}.  
We focus on stable 
extensions.%
\begin{definition}\label{def:ABAsemantics_stable}
	Let $F=(\A,\R)$ be an AF. A set $E\in \cf(F)$ is \emph{stable} ($E\in \stb(F)$) iff it attacks each 
    $x\in \A\setminus E$.
\end{definition}

\paragraph{Assumption-based Argumentation}\label{sec:ABA}
We assume a \emph{deductive system}, \ie\, a tuple $(\lit,\rules)$, where  
$\lit$ is a set of 
sentences and 
$\rules$ is a set of inference rules over $\lit$. 
A rule $r \in \rules$ has the form
$a_0 \leftarrow a_1,\ldots,a_n$, \suth $a_i \in \lit$ for all $0\leq i\leq n$; $head(r) := a_0$ is the \emph{head} and $body(r) := \{a_1,\ldots,a_n\}$ is the (possibly empty) \emph{body} of $r$.
\begin{definition}
	An ABA framework (ABAF)~\cite{BondarenkoDKT97} is a tuple $(\lit,\rules,\asm,\contraryempty)$, where $(\lit,\rules)$ is a deductive system, $\asm \subseteq \lit$ a (non-empty) set of assumptions, and  
	$\contraryempty:\asm\rightarrow \lit$ a contrary function.
\end{definition}
An ABAF 
$\D$ is \emph{flat} if $head(r)\notin \asm$ for all $r\in \rules$.
In this work we focus on \emph{flat} and \emph{finite} ABAFs, \ie\, $\lit$ and $\rules$ are finite. 
We also restrict attention to \lit\ consisting of atoms.

An atom $p \in \lit$ is \emph{tree-derivable}~\cite{DBLP:books/sp/09/DungKT09} from sets of assumptions 
$S \subseteq \asm$ and rules $R \subseteq \rules$, 
denoted by $S \vdash_R p$, if there is a finite rooted labeled tree $t$ s.t.\
i) the root of $t$ is labeled with $p$, 
ii) the set of labels for the leaves of $t$ is equal to $S$ or $S \cup \{\top\}$, and 
iii) for each node $v$ that is not a leaf of $t$ there is a rule $r\in R$ such that $v$ is labeled with $head(r)$ and labels of the children correspond to $body(r)$ or $\top$ if $body(r)=\emptyset$.
We write $S\vdash p$ iff there is  $R \subseteq \rules$ such that $S\vdash_R p$; $S\vdash p$ is called an \emph{ABA argument}. 

Let $S\subseteq \asm$.
By $\contrary{S}:=\{\contrary{a}\mid a\in S\}$  we denote the set of all contraries of $S$. 
$S$ \emph{attacks} 
$T\subseteq\asm$ if there are $S'\subseteq S$ and $a\in T$ s.t.\ $S' \vdash \contrary{a}$; if $S$ attacks $\{a\}$ we say $S$ attacks~$a$. 
$S$ is conflict-free ($S\in\cf(\D)$) if it does not attack itself. 
We recall the definition stable extensions for ABA. 
\begin{definition}\label{def:ABAsemantics standard semantics}
	Let  $\D=(\lit,\rules,\asm,\contraryempty)$ be an ABAF. 
    A set $S\subseteq \asm$ is a \emph{stable extension} ($S\in \stb(\D)$) iff $S$ is conflict-free and attacks each 
    $x\in \mathcal{A} \setminus S$.
\end{definition}

\begin{definition}\label{def:cred acc}
An assumption $a\in \asm$ is \emph{credulously accepted} \wrt\ stable extension semantics in an ABAF $\D=(\lit,\rules,\asm,\contraryempty)$ iff there is some $S\in\stb(\D)$ with $a\in S$;
$a$ is \emph{rejected} if it is not credulously accepted. 
\end{definition}

\begin{example}\label{ex:background}
Consider an ABAF 
$
\tuple{\lit,\rules,\asm,\contraryempty}$ with
$\allowbreak\lit \!=\! \{ a,b,c,d,p, \contraryc{a},\contraryc{b},\contraryc{c},\contraryc{d}\}$, 
$\asm \!=\! \{a,b,c,d\}$ with  contraries
 $\contraryc{a}$, $\contraryc{b}$, $\contraryc{c}$, $\contraryc{d}$, respectively, and rules $r_1,\dots, r_4$, respectively:
\begin{align*}
\contraryc{c}\la a,d &&p\la b  && \contraryc{d}\la c &&\contraryc{a}\la	 p,c&&
\end{align*}
We obtain two stable extensions:
 $\{b,c\}$ and $\{a,b,d\}$. Indeed, $\{b,c\}$ attacks $a$ and $d$ since $\{b,c\}\vdash \contrary{a}$ and $\{c\}\vdash\contrary{d}$; and $\{a,b,d\}$ attacks $c$ since $\{a,d\}\vdash \contrary{c}$.
\end{example}

Viewing tree derivations as arguments, an ABAF induces an AF as follows~\cite{CyrasFST2018}.
\begin{definition}
	The associated AF $F_\D=(\A,\R)$ of an ABA $\D\!=\!(\lit,\rules,\asm,\contraryempty)$
	is given by
	$A= \{ S\vdash p \mid \exists R\subseteq \mathcal R: S \vdash_R p \}$  
	and 
	$R$ such that $(S\vdash p,S'\vdash p')\in R$ iff $p\in \contrary{S'}$. 
\end{definition}

Semantics of AFs and ABAFs 
correspond as follows.

\begin{restatable}{proposition}{propSemanticsCorrespondenceStandardInst}
	\label{prop:semantics correspondence standard inst}
	Let $\D\!=\!(\lit,\rules,\asm,\contraryempty)$ be an ABAF and $F_{\D}$ its
    associated AF.
	If $E\!\in\! \stb(F_{\D})$ then $\bigcup_{S\vdash p\!\in E}S\!\in\! \stb(\D)$;
	if $S\!\in\! \stb(\D)$ then $\{S'\vdash p\mid \exists S'\subseteq S: S'\vdash p\}\in \stb(F_{\D})$.
\end{restatable}
Note that the AF instantiation can be exponential in the size of the given ABAF $\D$. To tackle this issue, ~\citeauthor{aba-to-af}~\shortcite{aba-to-af} propose a poly-time preprocessing technique that yields an ABAF so that the resulting AF is polynomial in the size of $\D$. Their procedure handles the derivation circularity and flattens out the nested argument construction;
they show that this construction preserves semantics under projection. 
In \S\ref{sec:eval} we use this procedure to construct the AF-based GNN baseline that we use to evaluate the performance of our native ABAF GNNs.

\subsection{Graph Neural Networks}\label{sec:GNNs}

A neural network (NN) is a parametrised function that maps an input vector \(\mathbf{x}\in\mathbb{R}^n\) to an output through successive layers of neurons. We set the input neurons as \(\mathbf{h}^{(0)} = \mathbf{x}\) and, denoting by \(\mathbf{h}^{(l)}\) the row-vector of neurons at layer \(l\), each neuron at layer \(l\) computes
\[
  h_i^{(l)} \;=\; f\!\bigl((\mathbf{w}_i^{(l-1)})^\top \mathbf{h}^{(l-1)} + b_i^{(l-1)}\bigr),
\]
where \(\mathbf{w}_i^{(l)}\) and \(b_i^{(l)}\) are the learnable weight-vector and bias for neuron \(i\) in layer \(l\), and \(f\) is a (possibly non-linear) activation function~\cite{Bishop07}.  In supervised learning, the NN is trained by comparing predictions to ground-truth labels via a loss function (e.g.\ cross-entropy
), and parameters are optimised by back-propagation and gradient descent.

Graph neural networks (GNNs) generalise NNs to consume graph‐structured data $G=(V,E)$, by jointly leveraging a node‐feature matrix $\mathbf{X}\in\mathbb{R}^{|V|\times d}$ (for $d$ features) and an adjacency matrix $\mathbf{A}$. Through repeated neighbourhood‐aggregation steps—each propagating (also known as \textit{message-passing}) and combining information from a node’s 
neighbours—a GNN learns \textit{embeddings} that encode both local topology and node attributes~\cite{bronstein2017geometric}. These embeddings can serve tasks such as node classification, where a final projection, a sigmoid and a threshold
yield per‐node labels. Popular aggregation schemes include Graph Convolutional Networks~\cite[\textbf{GCNs}]{GCN} and Graph Attention Networks~\cite[\textbf{GATs}]{graph-attention}.
\paragraph{GCNs} perform matrix‐based updates across all nodes simultaneously and treat every neighbour equally. 
GCNs update node representations by aggregating feature information from each node and its neighbours. At each layer $l$, the representation matrix $\mathbf{H}^{(l)}$ is computed from the previous layer $\mathbf{H}^{(l-1)}$ and the graph structure as:
\[
    \mathbf{H}^{(l)} = \sigma\left(\mathbf{\Delta}^{-1/2} \tilde{\mathbf{A}} \mathbf{\Delta}^{-1/2} \mathbf{H}^{(l-1)} \mathbf{W}^{(l-1)}\right),
\]
where $\tilde{\mathbf{A}} = \mathbf{A} + \mathbf{I}$ is the adjacency matrix of the graph with added self-loops (so each node includes its own features in the aggregation), $\mathbf{\Delta}$ is the diagonal degree matrix of $\tilde{\mathbf{A}}$, $\mathbf{W}^{(l)}$ is a learnable weight matrix for layer $l$, and $\sigma$ is a (possibly non-linear) activation function. The inclusion of self-loops ensures that nodes can retain and refine their own features across layers, rather than relying solely on neighbouring information. Typically, $\mathbf{H}^{(0)}=\mathbf{X}$.
\paragraph{GATs} compute learned attention coefficients~\cite{self-attention} to weight each neighbour’s contribution differently. 
Unlike GCNs, which treat all neighbouring nodes equally during aggregation, GATs introduce learnable attention mechanisms to assign different weights to different neighbours. Specifically, for a node $i$, attention coefficients $\alpha_{ij}$ are computed between $i$ and each of its neighbours $j$ using shared attention:
\[
    \alpha_{ij} = \frac{\exp\left(\text{LeakyReLU}\left(\mathbf{a}^\top [\mathbf{W} \mathbf{h}_i \,\Vert\, \mathbf{W} \mathbf{h}_j]\right)\right)}{\sum_{k \in \mathcal{N}(i)} \exp\left(\text{LeakyReLU}\left(\mathbf{a}^\top [\mathbf{W} \mathbf{h}_i \,\Vert\, \mathbf{W} \mathbf{h}_k]\right)\right)},
\]
where $\mathbf{W}$ is a learnable weight matrix, $\mathbf{a}$ is a learnable attention vector, $[\cdot\,\Vert\,\cdot]$ denotes vector concatenation, and $\mathcal{N}(i)$ is the set of neighbours of node~$i$. $\text{LeakyReLU}$ is a non-linear activation function that allows a small negative slope (typically $\epsilon = 0.2$) for negative inputs, helping to avoid dead neurons and improve gradient flow.
The updated representation of node $i$ is then computed as a weighted sum of the transformed features of its neighbours:
\(
    \mathbf{h}'_i = \sigma\left(\sum_{j \in \mathcal{N}(i)} \alpha_{ij} \mathbf{W} \mathbf{h}_j\right)
\).
To stabilise the learning process, GATv2~\cite{brody2022gatv2} employs multi-head attention: multiple independent attention mechanisms are applied in parallel, and their outputs are concatenated (in intermediate layers) or averaged (in the output layer).

When modelling heterogeneous graphs—where edges carry distinct semantics—one typically employs relation‐specific transformations, e.g.\ via a Heterogeneous Graph Convolution (HGC) module~\cite{heterogeneous_graphs_new}, and adds residual connections and normalisation between layers to stabilise training. 
To prevent overfitting, dropout~\cite{Dropout} can be employed after each layer, randomly setting to zero a subset of the weights, and early stopping to terminate training when the validation loss does not improve for a number of steps greater than a patience parameter.

\section{Predicting Acceptance with GNNs}\label{sec:GNN_model}

Here we detail our proposed GNN architecture to predict credulous acceptance of assumptions in an ABA framework. The core and novel component is the \emph{dependency graph}, a faithful graph-based representation of an ABAF, that guides the GNN learning of the ABA features that help predicting credulous acceptance 
(\S\ref{sec:dep_g}). 
The neural machinery adopted to learn to classify the assumption nodes of the dependency graph as \emph{credulously accepted} 
or \emph{rejected}  w.r.t.\ stable extension semantics is then detailed in \S\ref{sec:neural_arch}.

\subsection{Dependency Graph}\label{sec:dep_g}
Our proposed graph representation comprises nodes for each atom type (assumption and non-assumption) and for each rule, and three edge types: \emph{support} edges ($+$) linking body atoms to rule nodes; \emph{derive} edges ($\rhd$) linking rule nodes to their head elements; and \emph{attack} edges ($-$) connecting contraries to their assumption nodes.

\begin{definition}\label{def:dependency graph}
Let $\D=(\mathcal{L},\mathcal{R},\mathcal{A},\contraryempty)$ be an ABAF.
The \emph{dependency graph} $\dependency_D = (V,E,l)$ of $D$ is a directed, edge-labelled graph with nodes $V=\mathcal{L}\cup \mathcal{R}$, edges 
\begin{align*}
    E\ = \{(p,r)\mid r\in\mathcal{R}, p\in \body(r)\} \cup \{(r,p)\mid r\in\mathcal{R},\\ p\in \head(r)\} \cup \{(p,a)\mid a\in\mathcal{A}, \contrary{a}=p\},
\end{align*}
and edge labellings, for $e\in E$,
$$
l(e)=
\begin{cases}
    +,& \text{ if } e=(p,r), r\in\mathcal{R}, p\in \body(r)\\
    \rhd,& \text{ if } e=(r,p), r\in\mathcal{R}, p\in \head(r)\\
    -,& \text{ if } e=(p,a), a\in\mathcal{A}, \contrary{a}=p
\end{cases}$$
\end{definition}
\begin{observation}
$\dependency_{\D}\neq \dependency_{\D'}$ for every two ABAFs $\D,\D'$, $\D\neq\D'$,
i.e., $\dependency_{\D}$ is unique for each ABAF $\D$.
\end{observation}
Since $\dependency_{\D}$ contains information about all assumptions, contraries, and rules of a given ABAF $\D
$, the ABAF can be fully recovered from $\dependency_{\D}$, thus the semantics are preserved, as for the AF instantiation. 
In contrast to the AF instantiation, the construction of the dependency graph requires a single pass over the atoms and rules of $\D$.  

\begin{example}\label{ex:dep graph}
The dependency graph $\dependency_{\D}$ of the ABAF $\D$ from Example \ref{ex:background} is as follows:
\begin{center}
		\begin{tikzpicture}[xscale=1.4, yscale=1.1,>=stealth]
      
    {\small
    \node[
        anchor=north west,
        align=left,
        draw=black,
        fill=white,
        rounded corners=2pt,
        inner sep=3pt
    ] at (-0.5,2.3) {
        \textcolor{AssumpBlue}{\textbf{Assumption node}}\\
        \textcolor{RulePurple}{\textbf{Rule node}}\\
        $+$: support\\ $-$: attack\\ $\rhd$: derive
    };
    }
			\path 
			(-1,1)node[arg] (nb){$\contrary{b}$}
			(-1,0)node[aarg] (b){$b$}
            (0,0)node[rarg] (rr){$r_2$}
            (1,0)node[arg] (p){$p$}
            (2,0)node[rarg] (rrrr){$r_4$}
            (3,0)node[arg] (na){$\contrary{a}$}
            (4,0) node[aarg] (a){$a$}
            (4,1)node[rarg] (r){$r_1$}
            (3,1)node[arg] (nc){$\contrary{c}$}
		(2,1)node[aarg] (c){$c$}
            (2,2)node[rarg] (rrr){$r_3$}
            (3,2)node[arg] (nd){$\contrary{d}$}
		(4,2)node[aarg] (d){$d$}
            ;
            \path[->,>=stealth]
			(nb) edge node[left] {$-$}  (b)
            (b) edge node[above] {$+$}  (rr)
            (rr) edge node[above] {$\rhd$}  (p)
            (p) edge node[above] {$+$}  (rrrr)
            (c) edge node[left] {$+$}  (rrrr)
            (c) edge node[left] {$+$}  (rrr)
            (rrrr) edge node[above] {$\rhd$}  (na)
            (na) edge node[above] {$-$}  (a)
            (a) edge node[right] {$+$}  (r)
            (r) edge node[above] {$\rhd$}  (nc)
            (nc) edge node[above] {$-$}  (c)
            (rrr) edge node[above] {$\rhd$} (nd)
            (nd) edge node[above] {$-$} (d)
            (d) edge node[right] {$\rhd$} (r)
            ;
			
		\end{tikzpicture}
\end{center}
We can use $\dependency_{\D}$ to check that $b$ and $c$ jointly derive $\contrary{a}$, using rules $r_2$ and $r_4$.
\end{example}
\citeauthor{dependency-graph}~\shortcite{dependency-graph} define a dependency graph for ABA in the spirit of dependency graphs for logic programming~\cite{KonczakLS06,FandinnoL23}. However, their representation does not fully preserve the structure of the ABAF, preventing the possibility to extract the ABAF's extensions from a given dependency graph.\footnote{
In brief, the dependency graph from~\cite{dependency-graph} does not include rule nodes; positive edges $(p,q)$ are introduced whenever $p$ is contained in some rule body of a rule with head $q$. For example, the rule $(\contrary{a}\gets p,c)$ from Example~\ref{ex:dep graph} would yield the same edges as having two rules $(\contrary{a}\gets p),(\contrary{a}\gets c)$. The introduction of rule nodes was inspired by the conjunction node representation in~\cite{Li2021grASP}.
}
In contrast, our dependency graph is unique for every ABAF and 
preserves the semantics.

\begin{figure*}[ht]
    \centering
    \includegraphics[width=1\linewidth]{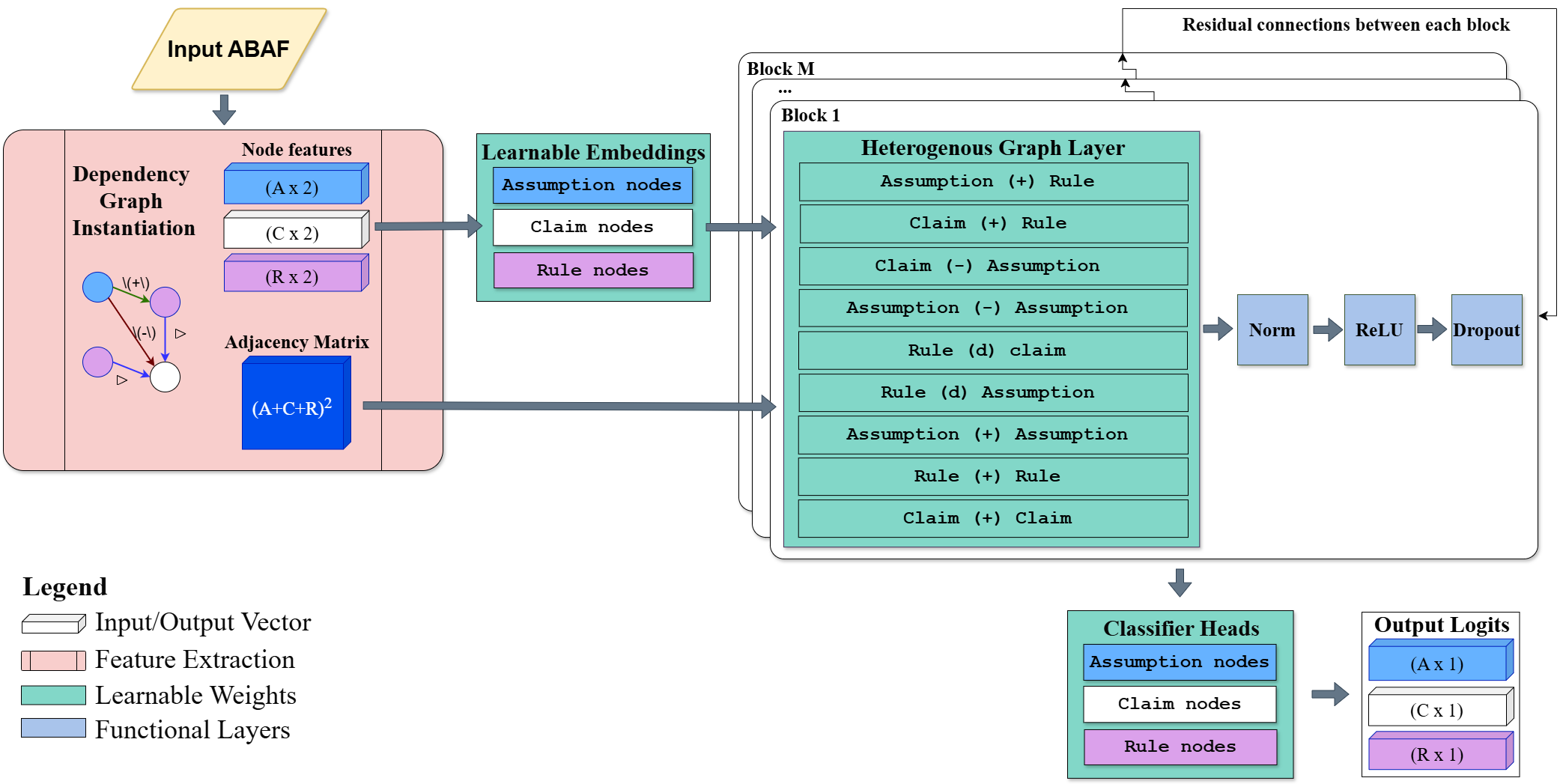}
    \caption{Diagram of the model architecture, including both the feature extraction (using the dependency graph described in \S\ref{sec:dep_g}) and the GNN described in \S\ref{sec:neural_arch}. The Heterogeneous Graph Layer module would apply Convolutional Layers for ABAGCN, or Attention Layers for ABAGAT.}
    \label{fig:final-model}
\end{figure*}
\subsection{Neural Architecture}\label{sec:neural_arch} 
Our GNN operates on the dependency graph’s adjacency matrix $\mathbf{W}\in\{0,1\}^{|V|\times|V|}$ and an initial node‐feature matrix $\mathbf{F}\in\mathbb{R}^{|V|\times 2}$, where each row $F_i$ contains the in‐ and out‐degrees of node $i$ for each node type: assumptions $\mathcal{A}$, non-assumptions (or claims) $\mathcal{C}=\mathcal{L\setminus A}$, and rules $\mathcal{R}$.  
Each node contains a self-loop of type '+' to propagate its features during learning. This is part of the GNN architecture (see  \S\ref{sec:GNN_model}) and does not interfere with the semantics; note that $\dependency_{\D}$ does not take edges $(s,t)\in\mathcal{L}^2$ into account. 
In addition, we maintain a learnable embedding matrix $\mathbf{L}\in\mathbb{R}^{|V|\times d_e}$ with embedding dimension $d_e$, whose entries are optimised jointly with the rest of the GNN. We show a diagram of our model architecture in Figure~\ref{fig:final-model}.
The core backbone of the GNN consists of $M$ blocks. In each block~$m$:
\begin{itemize}
  \item A \textsc{HGC} layer performs relation‐specific neighbourhood aggregation on the current embeddings $H^{(m-1)}$, using either GCN or GAT convolutional kernels (see \S\ref{sec:background}).
  \item The aggregated output is passed through functional layers for stability and regularisation: we use layer‐normalisation, ReLU activation, and dropout with rate $\delta$.
  \item A residual connection aggregates the block input $H^{(m-1)}$ to its output, yielding $H^{(m)}$.
\end{itemize}

This design allows distinct transformations per edge type while preserving gradient flow in deep architectures consisting of many layers. Each of these components (HGC and functional layers linked by residual connections) form a so called \textit{block}.
After $M$ blocks, with the parameter $M$ tuned on validation data, we extract the rows of $H^{(M)}$ corresponding to assumption, claim and rule nodes and apply a separate linear classifier head to produce logits $z\in\mathbb{R}^{A+C+R}$, with $A=|\mathcal{A}|, C=|\mathcal{C}|$, and $R=|\mathcal{R}|$. 

At inference time, a sigmoid activation yields the predicted probability of credulous acceptance for each assumption. Using a threshold $\tau$, again tuned on validation data, we classify assumptions  as accepted or rejected through a forward pass of the GNN that process the input features with our optimised weights obtained via supervised training.

We train the GNNs by minimising a weighted binary cross‐entropy loss on the assumption logits, using the Adam optimiser with learning rate $\lambda$.  Class weights compensate for the imbalance between accepted and rejected assumptions (see \S\ref{sec:data} and \ifappendix Table~\ref{tab:all-params} in Appendix~\ref{sec:data_det} \else Table 1 in Appendix B.1  of \cite{gehlot2025hggns}\fi for details on the training and test data).  

All hyperparameters—including the number of layer blocks $M$, the embedding dimension $d_e$, hidden dimension of each convolution layer, the dropout rate $\delta$, the learning rate $\lambda$, the batch size, class‐weight multiplier, and classification threshold $\tau$—are chosen via Bayesian optimisation~\cite{bayes} with 3‐fold cross‐validation, as implemented in \cite{wandb}. Details of tuning and hyperparameters are in \ifappendix Appendix~\ref{sec:tuning_det}\else Appendix B.3 of \cite{gehlot2025hggns}\fi.

\section{Extension Reconstruction}\label{sec:extension calc}
Here we outline our proposed extension-reconstruction algorithm, whose pseudo-code is shown in Algorithm~\ref{alg:enumerate-extension}.
Given an ABAF $\D$ and a predictor $\mathcal{M}$ in input, we pick the highest scoring assumption $a^*$ and mark it as  accepted (line 3). Before making our next guess, we modify $\D$ so that the extensions of the modified ABAF $\D'$ (under projection) correspond to the extensions of $\D$ containing $a^*$ (line 4). 
Since our prediction is imperfect, this modification may have introduced some inconsistencies in $\D'$, which we check in line~5.
If the check passes, we add $a^*$ to our extension $\mathcal{E}$.
\newcommand{\asa}{a^*}

We outline the $\textsc{modify}(\D,\asa)$ procedure for a given 
$\D$ and 
$\asa$.
When removing an assumption $a^*$ predicted accepted, we aim at ensuring that the extensions of the modified 
$\D'$ projected onto $\mathcal{A}$ correspond to the extensions of $\D$ that contain $a^*$.
We restrict to $\mathcal{A}$ because we introduce dummy-assumptions to ensure that rules that derive the contrary of $a^*$ act as constraints.
Concretely,  $\textsc{modify}(\D,a^*)$ performs the following steps:
\begin{enumerate}
\item We remove the assumption $a^*$ and all occurrences of it; that is, we let $\asm'=\asm\setminus \{\asa\}$ and we replace any rule $r$ with $r' = head(r)\gets  body(r)\setminus \{\asa\}$.
\item We modify each rule $r$ with $head(r)=\contrary{\asa}$: for each such rule $r$, we introduce a new dummy assumption $d_r$ and replace $r$ with  $r' = \contrary{d_r}\gets body(r)\cup d_r$. This modification ensures that not all body elements of $r$ can be true (accepted or derived) at the same time.
\item If $\asa$ is the contrary of another assumption $c$, then $c$ is rejected. We remove $c$ and any rule $r$ with $c\in body(r)$.
\item Lastly, we remove all facts from $\D$: for each rule $r$ with $body(r)=\emptyset$, we remove $r$ from 
$\rules$ and 
$p=head(r)$ from the body of each rule $r'$, i.e., we replace $r'$ with  $r'' = head(r')\gets  body(r')\setminus \{p\}$.
If $p$ is the contrary of an assumption $c$, the assumption is attacked and we proceed as in Step 3. We repeat this process until there are no remaining facts. 
\end{enumerate}
\begin{algorithm}[t]
\small
    \caption{Extension reconstruction sketch}
    \label{alg:enumerate-extension}
\begin{algorithmic}[1]
\REQUIRE ABAF $\D=(\mathcal{L},\mathcal{R},\mathcal{A},\contraryempty)$, predictor $\mathcal{M}$
\STATE Initialize extension $\mathcal{E} \leftarrow \emptyset$
\WHILE{$\mathcal{A} \neq \emptyset$}
    \STATE $a^* \leftarrow \arg\max_{a \in \mathcal{A}} P_{\mathcal{M}}(a)$ 
   \STATE $\D'\gets \textsc{modify}(\D, a^*)$   
    \IF{\textsc{isConflicting}($\D'$)}
        \STATE \textbf{break} 
    \ELSE
            \STATE $\mathcal{E} \leftarrow \mathcal{E} \cup \{a^*\}$ 
    \ENDIF
\ENDWHILE
\RETURN $\mathcal{E}$
\end{algorithmic}
\end{algorithm} 
Note that $\textsc{modify}(\D,a^*)$ runs in polynomial time; naively, the computation requires a loop over all rules in $\mathcal{R}$, for each $p\in \mathcal{L}$; thus, Algorithm~\ref{alg:enumerate-extension} is in $\P$. We establish the soundness of Algorithm~\ref{alg:enumerate-extension}.
\begin{restatable}
    {proposition}{propExtensionReconstruction}\label{prop:Ext rec}
    For an ABAF $\D$ and an assumption $\asa$, let $\D'$ denote the outcome of $\textsc{modify}(\D,\asa)$.
    If $\asa$ is credulously accepted in $\D$, then the following holds.
    \begin{align*}
\begin{split}
   \{S\!\cap\!\mathcal{A}\!\mid\!S\!\in\!\stb(\D')\}\!=\!\{S\!\setminus\!\{a^*\}\!\mid\!S\!\in\!\stb(\D),  a^*\!\in\!S \}
\end{split}
\end{align*}
\end{restatable}
The proof of Prop.~\ref{prop:Ext rec} is given in \ifappendix Appendix~\ref{app:extension calc}\else Appendix A of \cite{gehlot2025hggns}\fi.
With a perfect predictor of acceptance, Alg.~\ref{alg:enumerate-extension} would return a valid stable extension. With imperfect predictions, if $\asa$ is not accepted in $\D$, Alg.~\ref{alg:enumerate-extension} will encounter a rule of the form $\contrary{d_r}\gets d_r$ indicating a conflict.
\textsc{isConflicting}($\D'$) checks if such a rule exists 
(line 5). The partially computed set is returned if $\mathcal{A}=\emptyset$ or if a conflict is found. 

\section{Empirical Evaluation}\label{sec:eval}
In this section we assess the effectiveness of our proposed GNN models—both the convolutional (ABAGCN) and attention‐based (ABAGAT) variants—against our AFGCNv2-based baseline on two tasks over ABAFs: 
\begin{enumerate}
  \item \textbf{Credulous acceptance classification}: predicting, for each assumption, whether it is in any stable extension.  
  \item \textbf{Extension reconstruction}: recovering a full set of accepted assumptions (i.e.\ a predicted extension).
\end{enumerate}
For task 1 we report node‐level precision, recall, F1 and accuracy.  For task 2 we treat each stable extension as a set and measure performance using extension‐level F1 (details in \ifappendix Appendix~\ref{sec:metrics_det}\else Appendix B.2 of \cite{gehlot2025hggns}\fi).

\begin{figure*}[t]
  \centering
  \begin{subfigure}[b]{0.47\linewidth}
    \centering
    \includegraphics[width=\linewidth]{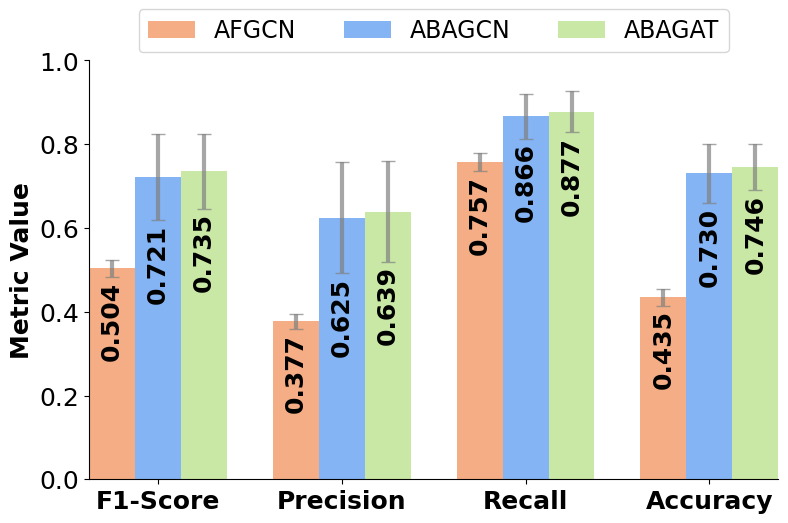}
    \caption{AF vs ABA GNNs on ICCMA data with $|\mathcal{L}|<100$}
    \label{fig:abafgcn_results}
  \end{subfigure}
  \hfill
  \begin{subfigure}[b]{0.47\linewidth}
    \centering
    \includegraphics[width=\linewidth]{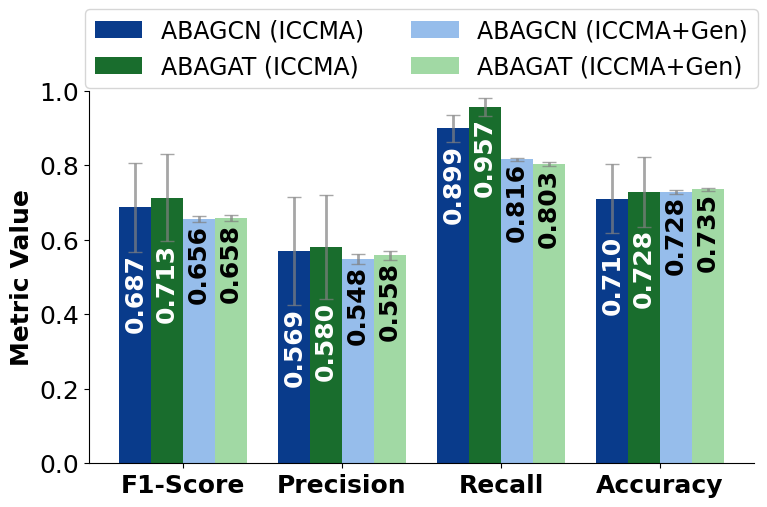}
    \caption{CGN vs GAT on ICCMA and generated data}
    \label{fig:abagAT_only}
  \end{subfigure}
  \caption{Model comparison according to F1, Precision, Recall and Accuracy on different cuts of data: small ICCMA ABAFs (with less than 100 elements) in panel (a) to be able to compare to the AFGCN baseline; Comparison between the full ICCMA test set and the test set augmented with our generated data in panel (b).}
  \label{fig:combined_abafgcn_abagat}
\end{figure*}

\subsection{Data}\label{sec:data}
ICCMA~2023 included an ABA track--but not an approximate ABA one--whose benchmark comprised of 400 ABAFs defined by four parameters: number of atoms, assumption proportion, maximum rules per atom, and maximum rule‐body length. Ground‐truth labels are obtained by running ASPForABA\footnote{Available at \url{https://bitbucket.org/coreo-group/aspforaba}.}~\cite{abaf_generator}
on every test instance with timeout threshold 10\,min, yielding 380 benchmark ABAFs with exact annotations within the ICCMA data. 

To obtain a larger training corpus, we used the ABAF generator from~\cite{abaf_generator} to produce 48,000 flat ABAFs (47,794 after timeout), extending the parameter ranges. Parameters are shown in \ifappendix Table~\ref{tab:all-params} in Appendix~\ref{sec:data_det}\else Table 1 in Appendix B.1 of \cite{gehlot2025hggns}\fi. To match the original assumptions acceptance rate (36.7\% overall, 60.3\% within ABAFs with at least one stable extension), we curated a subset of 19,500 ABAFs (13,000 with at least one accepted assumption, 6,500 with none), yielding 32.5\% overall acceptance and 53.4\% within ABAFs with at least a solution. 

Finally, we used a 75:25 train–test split, stratified across three groups—small ICCMA (25–100 atoms), full ICCMA, and our generated dataset—to ensure a fair comparison with AFGCNv2, which cannot process ABAFs larger than 100 atoms. Accordingly, all results are reported on three test sets: ICCMA small, ICCMA full and ICCMA+Generated.
\subsection{Baseline}\label{sec:abafgcn}
Our baseline implements a three‐stage pipeline that reduces ABAF credulous acceptance 
to an AF classification task:
\begin{enumerate}
    \item \textbf{ABAF-AF translation:}  Given an input ABAF, we use the method from \cite{aba-to-af} and implemented in the AcBar toolkit\footnote{Available at \url{https://bitbucket.org/lehtonen/acbar}}~\cite{lehtonen2021acbar} to convert the input ABAF into a (poly-sized) 
    AF in polynomial time.
    \item \textbf{Argument classification via AFGCNv2:}  The generated AF is represented by its adjacency matrix and graph‐level features, then fed into the AFGCNv2 solver (see \S\ref{sec:related}) using pre‐trained weights for credulous acceptance under stable semantics.  To ensure a fair comparison with our models, we disable its grounded‐extension pre‐check and apply a classification threshold—tuned on a held‐out validation set—to the sigmoid output so that any argument with score above the threshold is 
    accepted.
    \item \textbf{Element acceptability recovery:}  Finally, we derive assumption acceptance in the original ABAF: an assumption $a$ is accepted if the argument $\{a\}\vdash a$ is accepted.
\end{enumerate}
This baseline 
leverages an existing AF‐based GNN solver, providing a direct 
comparison for our native ABAF GNNs. 
\subsection{Credulous Acceptance Classification}
We first evaluate how well each model predicts assumption‐level acceptance under stable semantics. 
Figure~\ref{fig:abafgcn_results} shows that both our GNNs substantially outperform the AF baseline on the small ICCMA dataset ($|\mathcal{L}|<100$). 
We test on this set because AFGCNv2 cannot handle ABAFs greater than this size.
F1 increases from 0.5 (AFGCN) to 0.72 (ABAGCN) and 0.74 (ABAGAT); recall climbs by about 10 points; precision improves by about 25 points; and accuracy rises by over 30 points. ABAGAT achieves the highest results across all metrics on this test set, but it is not significantly different from ABAGCN. Statistical tests 
are provided in \ifappendix Table~\ref{tab:t_tests} in Appendix~\ref{sec:test_det}\else Table~3 in Appendix~B.5 of \cite{gehlot2025hggns}\fi.

In Figure~\ref{fig:abagAT_only} we examine the effect of testing only on the ICCMA data versus adding our generated instances. Both ABAGCN and ABAGAT see modest gains in accuracy when testing on the generated ABAFs together with the ICCMA ones, but incur a notable drop in precision and recall, resulting in a decrease in F1. This pattern indicates that synthetic augmentation possibly produced more challenging ABAFs than the ones in the competition.  Crucially, despite this worsening in performance, both GNN variants maintain strong performance across the held‐out ICCMA and generated test split, demonstrating robust generalisation to larger and more diverse ABA frameworks. Here ABAGCN and ABAGAT do not show significant difference across metrics, apart from ABAGAT surpassing ABAGCN in accuracy and recall on the ICCMA test, while ABAGCN showing significantly better recall than ABAGAT on the ICCMA+Gen set.

Comparing the small ICCMA results in Figure~\ref{fig:abafgcn_results} to the (full) ICCMA bars in Figure~\ref{fig:abagAT_only}, F1 dips by under one percentage point, precision by roughly two points and accuracy by about one point on the full dataset, while recall actually climbs by two, to six points, for ABAGCN and ABAGAT, respectively.  In other words, on larger ABAFs both GNNs trade a bit of overall and positive‐class fidelity for stronger coverage of accepted assumptions.
   
\subsection{Extension Reconstruction}\label{sec:extension calc exper}
For our second experiment, we test the poly-time algorithm from \S\ref{sec:extension calc}, using our approximate ABA models to calculate a stable extension.
Our constructive extension-reconstruction method achieves very high F1 (0.85) on small ABAFs (10 atoms), but performance drops sharply to around 0.60 by 50–100 atoms and then plateaus at approximately 0.58 for larger frameworks (see \ifappendix Figure~\ref{fig:enumerate-extension} in Appendix~\ref{sec:additional_exp}\else Figure~3 in Appendix~B.4 of \cite{gehlot2025hggns}\fi).
This decline reflects how early misclassifications by the GNN propagate through subsequent modification steps, reducing both precision and recall, 
but highlights opportunities for future work 
to mitigate degradation.

We also compare runtimes on 15 of the most challenging ABAFs (4,000–5,000 atoms) from our dataset, held out in the test set. The exact ASPForABA solver averaged 435s per instance, whereas our approximate extension reconstruction ran in 192s—2.3$\times$ faster—while still achieving F1 of 0.68 (0.77 accuracy).
These results highlight the potential of GNN‐driven approximate reasoning to yield speed-ups over exact methods, maintaining a similar level predictive performance as in the acceptance classification task.

\section{Conclusion}

We have demonstrated that heterogeneous GNNs over our dependency‐graph encoding can accurately and efficiently approximate credulous acceptance in ABA under stable semantics.  Both ABAGCN and ABAGAT outperform an AF‐based GNN baseline—achieving node‐level F1 of 0.65 overall and up to 0.71 on small frameworks—while our poly‐time extension‐reconstruction procedure reconstructs stable sets with F1 greater than 0.85 on small ABAFs and maintains F1 of about 0.58 at scales of 1,000 atoms.  Crucially, on the 4,000–5,000‐atom ABAFs where ASPForABA average runtime is 435s, our approximate extension are derived in 192s (2.3× faster) with F1 of 0.68, underscoring substantial runtime gains without sacrificing predictive quality.  These findings show that GNNs can bridge the gap between accuracy and tractability in structured argumentation.

Future work includes integrating lightweight symbolic checks to boost precision (e.g.\ grounded semantics as done in~\cite{afgcnv2}), extending the approach to other semantics, including admissible, complete, and preferred, where GNNs may be able to exploit local structural patterns even more efficiently. Additionally we aim at lifting the flatness restriction to handle general ABAFs and cover applications where non-flat ABAFs are used~\cite{causal}. Together, these directions promise more scalable and interpretable reasoning tools for complex argumentative scenarios. 

\section*{Acknowledgements}
Rapberger and Russo were funded by the ERC under the ERC-POC programme (grant number  101189053) while Toni under the EU’s Horizon 2020 research and innovation programme (grant number 101020934); Toni also by J.P. Morgan and by the Royal Academy of Engineering under the Research Chairs and Senior Research Fellowships scheme.

\ifappendix \appendix \else \end{document} \fi
\section{Correctness of the Extension Reconstruction Algorithm}\label{app:extension calc}
In this section, we prove the following proposition.
\propExtensionReconstruction*
To do so, we prove that
\begin{enumerate}
    \item a correctly guessed accepted assumption can be faithfully removed (correctness of steps 1-3), shown in Proposition~\ref{prop:correctness step 1-3}, and
    \item facts can be faithfully removed (correctness of step 4), shown in Proposition~\ref{prop:remove fact}.
\end{enumerate}

We introduce the following two operations. The first one removes facts from an ABAF, the second one removes rejected assumptions (used in Steps 3 and 4). 
\begin{definition}
Let  $\D=(\mathcal{L},\mathcal{R},\mathcal{A},\contraryempty)$ be an ABAF.
     We define the fact-removal operation $FR(\cdot,\cdot)$ that takes an ABAF $\D=(\mathcal{L},\mathcal{R},\mathcal{A},\contraryempty)$ and a fact $r=(p\gets)$ returns $FR(\D,r)=(\mathcal{L},\mathcal{R}^{rm_1},\mathcal{A},\contraryempty)$
     with $$\rules^{rm_1}=\{head(r')\gets  body(r')\setminus \{p\}\mid r'\in \rules,r'\neq r\}.$$
     Further, we define the assumption-removal operation $AR(\cdot,\cdot)$ that takes an ABAF $\D$ and an assumption $b$ and returns $AR(\D,b)=(\mathcal{L},\mathcal{R}^{rm_2},\mathcal{A}^{rm_2},\contraryempty)$
     with $\mathcal{A}^{rm_2}=\asm\setminus \{b\}$
     and $$\rules^{rm_2}=\{r\in \rules\mid b\notin  body(r)\}.$$
 \end{definition}
 We extend these operations to a set of rules/assumptions by iteratively applying the operations to each element in the set. 

We start by proving 2.\ and show that facts can be faithfully removed (step 4).
\begin{proposition}\label{prop:remove fact}
    Let  $\D=(\mathcal{L},\mathcal{R},\mathcal{A},\contraryempty)$ be an ABAF and let $r=(p\gets)\in\rules$. 
    Let $C=\{a\in \asm\mid \contrary{a}=p\}$.
    Then $\stb(\D)=\stb(FR(AR(\D,C),r))$.
\end{proposition}
\begin{proof}
Let $\D'=FR(AR(\D,C),r)$ with assumption set $\asm'$ and rule set $\rules'$.

First note that there is a 1-1 correspondence between arguments constructed from $S$ in $\D$ and $D'$, for each set $S$ of assumptions which does not contain elements from $C$. 
That is, for each $S\subseteq \asm$ with $S\cap C=\emptyset$,
$S\subseteq q$ is an argument in $\D$ iff $S\vdash q$ is an argument in $\D'$:

($\Rightarrow$) Let $S\vdash q$ be an argument in $\D$. Let $R\subseteq \rules$ such that $S\vdash_R q$ in $\D$.
By assumption, $b\notin body(r')$ for each $b\in C$ and for each $r'\in R$, thus, $r'$ is not deleted in $\rules'$.
In case that $r\in R$, 
then $r$ is `cut out' from the tree derivation in the rule modification: each occurrence from $p$ is removed from the body of each rule, and $r$ is removed. 
Let $R'$ denote the modified rules. 
We can construct an argument $S'\vdash_{R'\setminus \{r\}} \contrary{c}$.

($\Leftarrow$) For the other direction, $S\vdash q$ denote an argument in $\D'$ and let $R\subseteq \rules'$ denote a rule set such that $S\vdash_R q$ in $\D'$.
Let $R^\circ$ denote the set of rules corresponding to $R$ in $\D$.
In case $p\notin body(r')$ for each rule $r\in R^\circ$ then $R=R^\circ$ and $S\vdash_R q$ is a tree-derivation in $\D$. Otherwise, $S\vdash_{R^\circ\cup \{r\}} q$ is a tree-derivation in $\D$. 
In both cases, we obtain that $S\vdash q$ is an argument in $\D$. 
\begin{itemize}
    \item 
    First, let $S\in\stb(\D)$. 
    Then $b\notin S$ for each $b\in C$ since each $b$ is attacked by a fact.
    Thus, as shown above, we can construct the same arguments from $S$ in $\D$ and $D'$.
    Consequently, $S$ is conflict-free in $\D'$ (otherwise, $S$ would attack itself in $\D$ as well) and  attacks all assumptions in $\asm'\setminus S$ (since $\asm'\subseteq \asm$). We obtain $S\in \stb(\D')$.

    \item Now, let $S\in \stb(\D')$.
    $S$ does not contain arguments from $C$. Then $S$ is conflict-free in $\D$ since there is a 1-1 correspondence between the arguments constructed from $S$ in $\D$ and $\D'$ (thus, if there is a conflict in $S$ in $\D$ then $S$ attacks itself in $\D$' as well). Moreover,
    $S$ attacks each assumption in $\asm'\setminus S$ in $D$ by the same reasoning and since $S$ is stable in $\D'$.
    Each remaining assumption is attacked by the fact $r=(p\gets)$. Thus, $S$ is stable in $\D$.\qedhere    
\end{itemize}
\end{proof}

This shows that a fact can be removed without altering the stable sets.
Iterative application of Proposition~\ref{prop:remove fact} shows correctness of step 4.

Next we show 2. and prove that that steps 1-3 are correct. 
We define the ABAF modification \emph{accept assumption} $AA(\D,a)$ below. 
 \begin{definition}
     Let  $\D=(\mathcal{L},\mathcal{R},\mathcal{A},\contraryempty)$ be an ABAF, let $a\in \mathcal{A}$, let $C=\{c\in\asm\mid \contrary{c}=a\}$ and let $R_{\contrary{a}}=\{r\in\rules\mid head(r)=\contrary{a}\}$.
     We define $AA(\D,a)=(\mathcal{L},\mathcal{R}',\mathcal{A}',\contraryempty)$ with 
     $$\asm'=(\asm\cup \{d_r\mid r\in R_{\contrary{a}}\}) \setminus (C\cup \{a\})$$ and 
     \begin{align*}
        & R'=\\&\! \{head(r)\!\gets\!  body(r)\!\setminus\! \{a\}\!\mid\! r\!\in\! \mathcal{R}\!\setminus\! R_{\contrary{a}},C\cap body(r)\!=\!\emptyset\}\cup \\
         & \ \{\contrary{d_r}\!\gets\!  (body(r)\cup d_r)\!\setminus\! \{a\}\mid r\in R_{\contrary{a}},C\cap body(r)\!=\!\emptyset\}.
     \end{align*}
 \end{definition}
We show that this ABAF satisfies the equation in Proposition~\ref{prop:Ext rec}.
\begin{proposition}\label{prop:correctness step 1-3}
 Let  $\D=(\mathcal{L},\mathcal{R},\mathcal{A},\contraryempty)$ be an ABAF, let $a\in \mathcal{A}$ be credulously accepted with respect to stable semantics in $\D$, and let $\D'=AA(\D,a)$. Then the following is satisfied. 
    \begin{align*}
   \{S\cap \mathcal{A} \mid S\in \stb(\D')\}=  \{ S \setminus \{a\} \mid S \in \stb(\D),\ a \in S \}.
\end{align*}
\end{proposition}
\begin{proof} Let $C=\{c\in\asm\mid \contrary{c}=a\}$ denote all assumptions with $a$ as its contrary. Let $\D'=RAA(\D,a)=(\mathcal{L},\mathcal{R}',\mathcal{A}',\contraryempty)$.

\begin{table*}[t]
  \centering
  \caption{Parameter ranges for ICCMA benchmarks and additional synthetic data}
  \label{tab:all-params}
  \begin{tabular}{@{} m{5.5cm} m{3.5cm} m{4.5cm} @{}}
    \toprule
    \textbf{Parameter}                   & \textbf{ICCMA}                 & \textbf{Generated}          \\
    \midrule
    Acceptance Rate (with at least one ext)             & 36.7\% (60.3\%)                            & 32.5\% (53.4\%)      \\
    Number of atoms                      & 25, 100, 500, 2000, 5000              & 25, 50, 75, 100, 250, 500, 750, 1000, 2000, 3000, 4000, 5000 \\
    Assumption proportion                & 10\%, 30\%                            & 25\%, 40\%, 50\%, 60\%, 75\%      \\
    Max.\ rules per derived atom         & 5, 10                                 & 2, 4, 8, 16                       \\
    Max.\ rule-body length               & 5, 10                                 & 2, 4, 8, 16                       \\
    Cycle cap                            & \textemdash                          & 0.01, 0.03, 0.05, 0.07, 0.09, 1.0  \\
    Total instances (post-filtering)     & 400 (380)                                   & 48\,000 (19\,500) \\
    \bottomrule
  \end{tabular}
\end{table*}

By assumption, the ABAF $\D$ has a stable extension (since $a$ is credulously accepted). 
Moreover, by construction, each assumption of the form $d_r$ can be accepted in case $\D'$ has a stable extension. Thus, each stable extension contains all $\{d_r\mid r\in R_{\contrary{a}}\}$. 
    \begin{itemize}
        \item Let $S\in\stb(\D')$.
        We show that $S'=(S\cap \asm)\cup \{a\}\in \stb(\D)$.

        $S'$ is conflict-free in $\D$: towards a contradiction, suppose $S'$ is not conflict-free. 
        Let $T\vdash_R \contrary{b}$ for some $T\subseteq S'$, $R\subseteq \rules$, $b\in S'$, in $\D$.
        Note that $body(r)\cap C=\emptyset$ for each $r\in R$ since $C\cap S'=\emptyset$.
        We proceed by case distinction.

        \underline{Case 1: $head(r)\neq \contrary{a}$ for all $r\in R$.}
        Let $R'$ denote the modified rules in $\D'$ corresponding to $R$. Each rule $r\in R$ either stays unchanged (in case $a\notin body(r))$ or the assumption $a$ is removed from the body. 
        We can construct an argument in $\D'$ based on $T\setminus \{a\}$ using $R'$ that derives $\contrary{b}$. Thus, $b$ is attacked by $S$ in $\D'$, contradiction to $S$ being stable in $D'$.  

        \underline{Case 2: there is some $r\in R$ with $head(r)=\contrary{a}$.}
        Wlog, we can assume that the top-rule $r$ of the argument satisfies $head(r)=\contrary{a}$ (otherwise choose an appropriate sub-argument). 
        By definition, this induces a rule $\contrary{d_r}\gets  (body(r)\cup d_r)\setminus \{a\}$ in $\D'$. 
        Every element in $body(r)$ can be derived from $S$ in $\D'$ (this can be proven analogous to Case 1). 
        Thus, we can construct an argument from assumptions in $S$ (recall that $\{d_r\mid r\in R_{\contrary{a}}\}\subseteq S$) that derives $\contrary{d_r}$. Thus, $S$ is conflicting in $\D'$, contradiction to the assumption $S\in\stb(\D')$.

        We have shown that $S'$ is conflict-free in $\D$.
        It remains to show that $S'$ attacks each assumption in $\asm\setminus S'$ in $\D$.
        Let $b\in \asm\setminus S'$.
        The assumption $b$ is not contained in $S$ in $\D'$. 
        
        \underline{Case 1: $b\notin \asm'$.} Then $b\in C$. Since $a\in S'$, $b$ is attacked. 
        
        \underline{Case 2: $b\in \asm'$.} Then $b$ is attacked by $S$ in $\D'$. That is, there is 
        a tree-derivation $T\vdash_R \contrary{b}$ in $D'$ with $T\subseteq S$. 
        We observe that (i) $\contrary{a}$ does not appear in the tree derivation (otherwise, we can construct an argument with conclusion $\contrary{a}$ from assumptions in $S$ in $\D'$, contradiction to $S$ being stable in $\D'$); and (ii) assumptions of the form $d_r$ do not appear in the tree derivation (otherwise we can derive a contradiction to conflict-freeness of $S$ like in Case 2). 
        For each $r\in R$, either $r\in \rules$ or $(head(r)\gets body(r)\cup \{a\})\in \rules$.
        We thus can use the rules in $R$ to construct a tree-derivation from $T\cup \{a\}$ with conclusion $\contrary{b}$ in $\D$ and obtain $S'\in \stb(\D)$, as desired. 
        . 
        
        \item Let $S\in\stb(\D)$ with $a\in S$.
        Let $$S'=(S\setminus \{a\})\cup \{d_r\mid r\in R_{\contrary{a}}\}.$$
        We show that $S'\in\stb(\D')$.

        $S'$ is conflict-free: towards a contradiction, suppose there is a tree derivation $T\vdash_R \contrary{b}$ for some $b\in S'$ in $\D'$.

        \underline{Case 1: $b$ is of the form $d_r$ for some $r\in R_{\contrary{a}}$.}
        Wlog let $d_r$ be the only occurrence of assumptions of this form (otherwise, choose an appropriate sub-argument). Then each $p\in body(r)$ can be derived from assumptions in $S$. Thus, we can construct an argument with conclusion $\contrary{a}$ from $S$, contradiction to conflict-freeness of $S$ in $\D$.

        \underline{Case 2: $b\in \contrary{\asm}$.}
        The only possible rule modifications to obtain the rules in $R$ (from rules in $\rules$) is the removal of assumption $a$. 
        That is, for each rule $r\in R$, either $r\in \rules$ or $(head(r)\gets body(r)\cup \{a\})\in \rules$. Let $R'$ denote the rules corresponding to $R$ in $\D$. 
        We can construct an argument in $\D$ based on $T\cup \{a\}$ (in case $R\neq R'$) or $T$ (in case $R=R'$) that derives $\contrary{b}$. Thus, $b$ is attacked by $S$ in $\D$, contradiction to conflict-freeness of $S$ in $\D$.

        We have shown that $S'$ is conflict-free in $\D'$.
        It remains to show that $S'$ attacks all remaining assumptions in $\D'$.
        By definition of $S'$, all remaining assumptions are contained in $\asm$ as well.
        We thus can proceed analogous to Case 2 to construct arguments with the appropriate conclusions and conclude that each remaining assumption is attacked by $S'$.
        Thus, $S'$ is stable in $D'$.\qedhere
    \end{itemize}
\end{proof}
By Proposition~\ref{prop:correctness step 1-3}, and by iterative application of Proposition~\ref{prop:remove fact}, we obtain that the equation in Proposition~\ref{prop:Ext rec} is correct. This implies that the algorithm presented in \S\ref{sec:extension calc} successfully returns a stable extension, provided the predictions are correct. 

\section{Details on Experiments}
\label{sec:exp_det}

Here we provide additional details for the experiments presented in \S\ref{sec:eval}. We cover data, hyperparameter tuning and additional details about performance, including statistical tests for the observed differences amongst methods.

\begin{table*}[ht]
  \centering
  \caption{Tested hyperparameter ranges and best configurations for GCN and GAT}
  \label{tab:gcngat_hyperparams}
  \begin{tabular}{@{}l l c c@{}}
    \toprule
    \textbf{Parameter}                & \textbf{Values tested}               & \textbf{GCN (best)} & \textbf{GAT (best)} \\
    \midrule
    Validation F1 Score               & —                                    & 0.6682              & 0.6709              \\
    Embedding dimension               & 16, 32, 64, 128, 256                 & 32                  & 64                  \\
    Hidden dimension                  & 32, 64, 128, 256, 512                & 32                  & 64                  \\
    Number of layers blocks           & 2–10                                 & 10                  & 10                  \\
    Dropout rate                      & 0–0.5 (uniform)                      & 0.0294              & 0.2198              \\
    Learning rate                     & $10^{-2}$–$10^{-4}$ (uniform)        & 0.0086              & 0.0066              \\
    Batch size                        & 16, 32, 64, 128                      & 128                 & 64                  \\
    Class imbalance weight            & 1.25, 1.5, 1.75, 2.0, 2.25           & 2.25                & 1.75                \\
    Classification threshold          & 0.20–0.85 (step 0.05)                & 0.50                & 0.45                \\
    Early stopping patience           & 10, 30, 50, 70, 90                   & 10                  & 30                  \\
    \bottomrule
  \end{tabular}
\end{table*}
\subsection{Data}\label{sec:data_det}
We began with the 400 ABAFs from the ICCMA~2023 benchmark,\footnote{Available at \url{https://zenodo.org/records/8348039}} each parameterised by the number of atoms, the proportion of those atoms that are assumptions, the maximum number of rules per derived atom, and the maximum rule‐body length. After running ASPForABA with a 10\,min timeout per instance, we retained 380 ABAFs annotated with credulous accepted labels for each assumption. 

To augment this corpus, we used the ABAF generator in \cite{abaf_generator} to produce 48,000 additional “flat” frameworks, extending each of the four original parameters as shown in Table~\ref{tab:all-params}. We then filtered out any instances that timed out during labelling (yielding 47,794), and from these we sampled a curated subset of 19,500 ABAFs so as to closely match the acceptance rates observed in the ICCMA data (32.5\,\% overall, 53.4\,\% among solvable instances). In particular, we balanced the generated set to include 13,000 ABAFs with at least one accepted assumption and 6,500 with none. 
Finally, to support fair evaluation of both our GNN models and the AFGCNv2‐based baseline—which cannot predict acceptance for ABAFs with more than 100 atoms—we stratified the entire collection into three groups (small ICCMA: 25–100 atoms; full ICCMA; and Generated) and performed a 75/25 train–test split within each group. Our augmented dataset is provided in our repository.\footnote{Available at \url{https://github.com/briziorusso/GNN4ABA}}

\subsubsection{Computing infrastructure} 
All the results were ran on NVIDIA(R) GeForce RTX 4090 GPU with 24GB dedicated RAM.

\subsection{Evaluation Metrics}\label{sec:metrics_det}
We evaluated the classification results using four metrics:
\begin{itemize}
    \item Precision = TP/(TP + FP)
    \item Recall = TP/(TP + FN)
    \item F1 = $2\times$(Precision$\times$Recall)/(Precision+Recall)
    \item Accuracy = TP/(TP + FP + TN + FN)
\end{itemize}

Precision and Recall measure the proportion of correct classifications based on the predicted classes and the true ones, respectively. In the formulae, True Positive (TP) is the number of estimated assumptions correctly credulously accepted; False Positive (FP) is an assumption which is credulously accepted but is not in any extension. True Negative (TN) and False Negative (FN) are assumptions that are not in any extension and are correctly (resp, incorrectly) classified as not accepted. F1 score is the harmonic mean of precision and recall.  
Finally, accuracy just looks at the TP rate over the full assumption set. 
For the credulously acceptance classification the metrics are calculated over the full set of assumption, while for the extension-reconstruction task they assumption set is restricted to the assumptions in any of the valid extensions.

\subsection{Hyperparameter Tuning}\label{sec:tuning_det}
We conducted hyperparameter optimisation separately for the ABAGCN and ABAGAT models using Bayesian search over the ranges reported in Table~\ref{tab:gcngat_hyperparams}. For each candidate configuration we trained on the 90\,\% of the 75\,\% training split and selected the model checkpoint with the highest F1 on the 10\,\% held‐out validation set, employing early stopping with patience drawn from \{10,\,30,\,50,\,70,\,90\}. Our search revealed that the attention‐based variant (ABAGAT) attained a marginally higher peak F1 (0.6709 vs.\ 0.6682) by leveraging a larger embedding and hidden dimension (64 vs.\ 32), higher dropout ($\simeq$0.22 vs.\ 0.03), and a smaller batch size (64 vs.\ 128). Interestingly, both models benefited from deeper architectures—ten message‐passing layers—as well as non‐trivial class‐imbalance weighting (1.75-2.25) and mid‐range learning rates ($6-9e^{-3}$). We also observed that the attention mechanism appeared to regularize the model, allowing for higher dropout and greater early‐stopping patience without degrading validation performance. 

\begin{figure}[!b]
\centering
\includegraphics[width=1\linewidth]{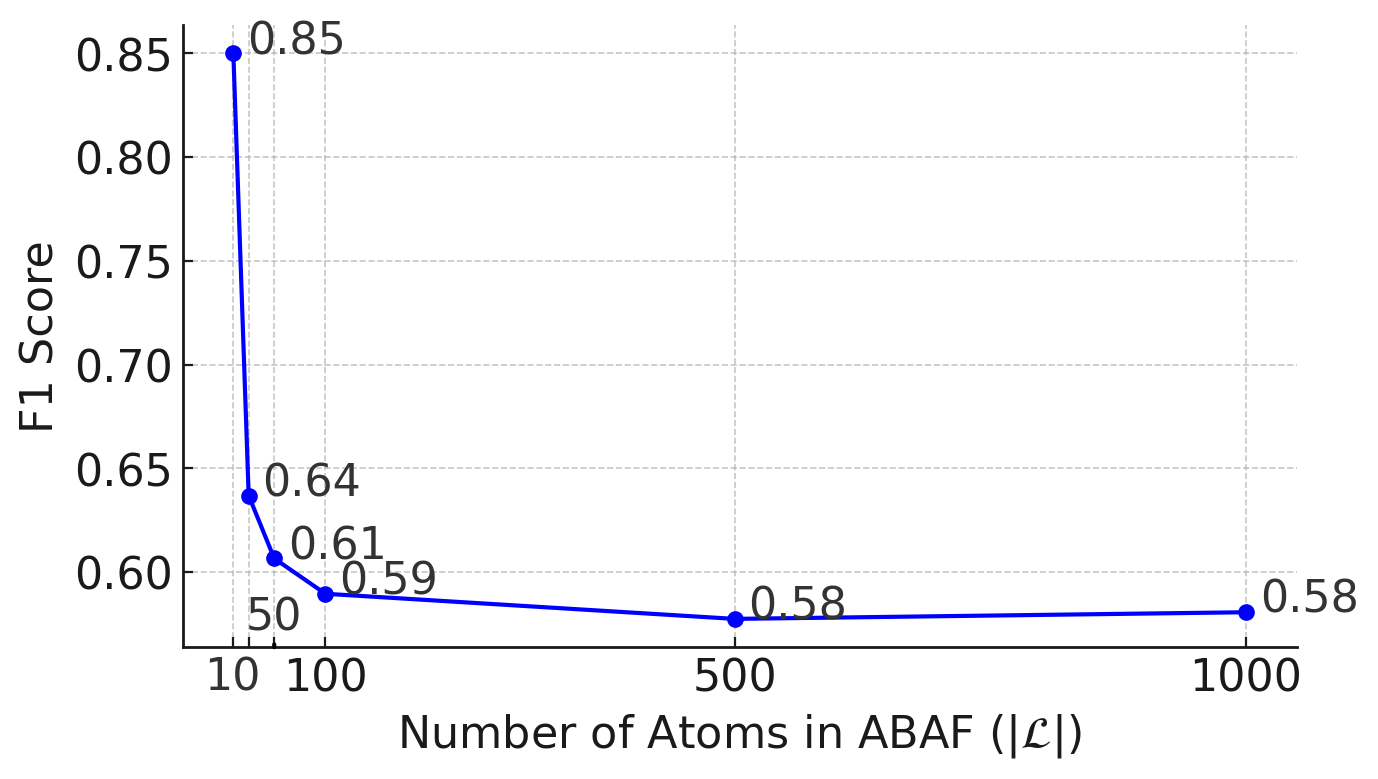}
\caption{F1 score of the extension by ABAF size.}
\label{fig:enumerate-extension}
\end{figure}
\subsection{Extensions Accuracy}\label{sec:additional_exp}
Beyond node‐level classification, we evaluated how well our extension‐reconstruction algorithm (\S\ref{sec:extension calc}) recovers full stable extensions. We plot the results discussed in the main text is Figure~\ref{fig:enumerate-extension} where we show the F1 of reconstructed extensions as a function of ABAF size. On very small frameworks (10 atoms), our method achieves high fidelity (F1$\sim$0.85), but performance degrades steadily as the number of atoms grows, falling to $\sim$0.60 for 50–100 atoms and plateauing around 0.58 for larger instances.
\begin{table*}[!t]
  \centering
  \begin{subtable}[t]{\textwidth}
    \centering
    \begin{tabular}{lcllll}
    \toprule
    \textbf{Scenario}   & \textbf{Model} & \textbf{F1 Score}     & \textbf{Accuracy}      & \textbf{Precision}     & \textbf{Recall}        \\
    \midrule
    \multirow{3}{*}{Small ICCMA} 
        & AFGCNv2 
            & 0.504$\pm$0.020 
            & 0.435$\pm$0.020 
            & 0.377$\pm$0.018 
            & 0.758$\pm$0.022 \\
        & ABAGCN 
            & \textbf{0.721$\pm$0.103}$^{***}$ 
            & \textbf{0.730$\pm$0.071}$^{***}$ 
            & \textbf{0.625$\pm$0.133}$^{***}$ 
            & \textbf{0.866$\pm$0.053}$^{***}$ \\
        & ABAGAT 
            & \textbf{0.735$\pm$0.089}$^{***}$ 
            & \textbf{0.746$\pm$0.055}$^{***}$ 
            & \textbf{0.639$\pm$0.120}$^{***}$ 
            & \textbf{0.877$\pm$0.049}$^{***}$ \\
    \midrule
    \multirow{2}{*}{All ICCMA}  
        & ABAGCN 
            & 0.687$\pm$0.119 
            & 0.710$\pm$0.093 
            & 0.569$\pm$0.145 
            & 0.899$\pm$0.037 \\
        & ABAGAT 
            & 0.713$\pm$0.116 
            & 0.728$\pm$0.094 
            & 0.580$\pm$0.141 
            & \textbf{0.957$\pm$0.024}$^{**}$ \\
    \midrule
    \multirow{2}{*}{ICCMA+Gen}  
        & ABAGCN 
            & 0.656$\pm$0.009 
            & 0.728$\pm$0.005 
            & 0.548$\pm$0.012 
            & \textbf{0.816$\pm$0.004}$^{***}$ \\
        & ABAGAT 
            & 0.658$\pm$0.009 
            & \textbf{0.735$\pm$0.004}$^{**}$ 
            & \textbf{0.558$\pm$0.012} \textbf{.} 
            & 0.803$\pm$0.005 \\
    \bottomrule
    \end{tabular}    
    \subcaption{Mean and Standard Deviation ($\mu\pm\sigma$)}
    \label{tab:iccma_all_scenarios_sig}
  \end{subtable}\hfill
  \vspace{0.5cm}
  \begin{subtable}[t]{\textwidth}
    \centering
\begin{tabular}{llllrl}
\toprule
Scenario   & Metric     & Method 1 & Method 2 & $t$      & $p$‑value \\
\midrule
Small      & F1         
    & ABAGAT (0.735 $\pm$ 0.089)  
    & ABAGCN (0.721 $\pm$ 0.103)  
    & 0.329    & 0.746       \\
Small      & F1         
    & ABAGAT (0.735 $\pm$ 0.089)  
    & AFGCNv2 (0.504 $\pm$ 0.020) 
    & 8.029    & 0.000***   \\
Small      & F1         
    & ABAGCN (0.721 $\pm$ 0.103)  
    & AFGCNv2 (0.504 $\pm$ 0.020) 
    & 6.579    & 0.000***   \\

Small      & Accuracy   
    & ABAGAT (0.746 $\pm$ 0.055)  
    & ABAGCN (0.730 $\pm$ 0.071)  
    & 0.557    & 0.585       \\
Small      & Accuracy   
    & ABAGAT (0.746 $\pm$ 0.055)  
    & AFGCNv2 (0.435 $\pm$ 0.020) 
    & 16.669   & 0.000***   \\
Small      & Accuracy   
    & ABAGCN (0.730 $\pm$ 0.071)  
    & AFGCNv2 (0.435 $\pm$ 0.020) 
    & 12.728   & 0.000***   \\

Small      & Precision  
    & ABAGAT (0.639 $\pm$ 0.120)  
    & ABAGCN (0.625 $\pm$ 0.133)  
    & 0.253    & 0.803       \\
Small      & Precision  
    & ABAGAT (0.639 $\pm$ 0.120)  
    & AFGCNv2 (0.377 $\pm$ 0.018) 
    & 6.839    & 0.000***   \\
Small      & Precision  
    & ABAGCN (0.625 $\pm$ 0.133)  
    & AFGCNv2 (0.377 $\pm$ 0.018) 
    & 5.849    & 0.000***   \\

Small      & Recall     
    & ABAGAT (0.877 $\pm$ 0.049)  
    & ABAGCN (0.866 $\pm$ 0.053)  
    & 0.482    & 0.636       \\
Small      & Recall     
    & ABAGAT (0.877 $\pm$ 0.049)  
    & AFGCNv2 (0.758 $\pm$ 0.022) 
    & 7.020    & 0.000***   \\
Small      & Recall     
    & ABAGCN (0.866 $\pm$ 0.053)  
    & AFGCNv2 (0.758 $\pm$ 0.022) 
    & 5.989    & 0.000***   \\

Full       & F1         
    & ABAGAT (0.713 $\pm$ 0.116)  
    & ABAGCN (0.687 $\pm$ 0.119)  
    & 0.493    & 0.628       \\

Full       & Accuracy   
    & ABAGAT (0.728 $\pm$ 0.094)  
    & ABAGCN (0.710 $\pm$ 0.093)  
    & 0.432    & 0.671       \\

Full       & Precision  
    & ABAGAT (0.580 $\pm$ 0.141)  
    & ABAGCN (0.569 $\pm$ 0.145)  
    & 0.183    & 0.857       \\

Full       & Recall     
    & ABAGAT (0.957 $\pm$ 0.024)  
    & ABAGCN (0.899 $\pm$ 0.037)  
    & 4.183    & 0.001**    \\

Full+Gen   & F1         
    & ABAGAT (0.658 $\pm$ 0.009)  
    & ABAGCN (0.656 $\pm$ 0.009)  
    & 0.703    & 0.491       \\

Full+Gen   & Accuracy   
    & ABAGAT (0.735 $\pm$ 0.004)  
    & ABAGCN (0.728 $\pm$ 0.005)  
    & 3.822    & 0.001**    \\

Full+Gen   & Precision  
    & ABAGAT (0.558 $\pm$ 0.012)  
    & ABAGCN (0.548 $\pm$ 0.012)  
    & 1.829    & 0.084 .     \\

Full+Gen   & Recall     
    & ABAGAT (0.803 $\pm$ 0.005)  
    & ABAGCN (0.816 $\pm$ 0.004)  
    & -6.500   & 0.000***   \\
\bottomrule
\end{tabular}    
\subcaption{Welch t‑test results for each metric and scenario.} \label{tab:t_tests_sub}
  \end{subtable}
  \caption{(a) Mean and Standard Deviation ($\mu\pm\sigma$) summary and (b) full t‑test breakdown for ICCMA small (25/100 element files), full (all element files), and ICCMA+Gen datasets. Sample sizes: 30/40 (small), 71/95 (full), 3727/4970 (ICCMA+Gen); Bootstrap samples: 10. Significance codes: 0‘***’0.001‘**’0.01‘*’0.05‘.’0.1.}\label{tab:t_tests}
\end{table*}
This trend reflects the compounding effect of early misclassifications: 
a single mistaken assumption in the initial GNN output can cascade through the constructive search steps and disproportionately harm both precision and recall. We envisage good opportunities to address this via lookback and lookahead error correction strategies to prevent error propagation.

\subsection{Statistical Tests}\label{sec:test_det}
To quantify the reliability of the performance differences among our models, we conducted Welch’s two‐sample \(t\)‑tests on each metric in each evaluation scenario. 

Table~\ref{tab:iccma_all_scenarios_sig} reproduces the mean (\(\mu\)) and standard deviation (\(\sigma\)) for F1, accuracy, precision, and recall in each scenario (Small ICCMA, All ICCMA, and ICCMA+Generated), exactly as plotted in Figure~\ref{fig:combined_abafgcn_abagat} of the main text. In this table each “\(\mu\pm\sigma\)” entry carries a superscript significance code—\texttt{***} for \(p<0.001\), \texttt{**} for \(p<0.01\), \texttt{*} for \(p<0.05\), and \texttt{.} for \(p<0.1\)—and is typeset in \textbf{bold} only when the top‐ranked model’s mean is significantly higher than the runner‐up at one of these levels.
Table~\ref{tab:t_tests_sub} provides the full pairwise comparison details that underpin the annotations in Table~\ref{tab:iccma_all_scenarios_sig}. For each scenario and metric it reports the two methods being compared, the Welch \(t\)‑statistic, the exact \(p\)‑value, and the same significance codes. This detailed breakdown allows readers to verify the strength and direction of every statistical difference that has been collapsed into the boldface annotations of the first table.

\end{document}